\documentclass{article}

\usepackage[a4paper, total={6in, 9.45in}]{geometry}

\usepackage[utf8]{inputenc}
\usepackage[T1]{fontenc} 
\usepackage{hyperref}      
\usepackage{url}     
\usepackage{booktabs}   
\usepackage{amsfonts}   
\usepackage{nicefrac}   
\usepackage{microtype} 
\usepackage{xcolor}    
\usepackage{graphicx}

\usepackage{booktabs}
\usepackage{amssymb}
\usepackage{amsmath}
\usepackage{bm}
\usepackage{subcaption}
\usepackage{algorithm}
\usepackage{algorithmic}
\usepackage{multirow}

\usepackage[colorinlistoftodos]{todonotes}

\usepackage{wrapfig}

\title{CHALLENGER: Training with Attribution Maps}

\date{}
\author{%
  Christian Tomani and Daniel Cremers \\ \\
  Technical University of Munich \\
  \texttt{\{christian.tomani, cremers\}@tum.de} \\
}

\begin{document}

\maketitle

\begin{abstract}

We show that utilizing attribution maps for training neural networks can improve regularization of models and thus increase performance. Regularization is key in deep learning, especially when training complex models on relatively small datasets. In order to understand inner workings of neural networks, attribution methods such as Layer-wise Relevance Propagation (LRP) have been extensively studied, particularly for interpreting the relevance of input features. We introduce \textit{Challenger}, a module that leverages the explainable power of attribution maps in order to manipulate particularly relevant input patterns. Therefore, exposing and subsequently resolving regions of ambiguity towards separating classes on the ground-truth data manifold, an issue that arises particularly when training models on rather small datasets. Our \textit{Challenger} module increases model performance through building more diverse filters within the network and can be applied to any input data domain. We demonstrate that our approach results in substantially better classification as well as calibration performance on datasets with only a few samples up to datasets with thousands of samples. In particular, we show that our generic domain-independent approach yields state-of-the-art results in vision, natural language processing and on time series tasks. 
  
\end{abstract}

\section{Introduction}

Deep neural networks have dramatically advanced the state-of-the-art in many different areas of machine learning. A driving force behind this development has been the drastic increase in model parameters. On the one hand this fact leads to highly expressive models that can learn complicated non-linear relationships from data, yet on the other hand it comes with a price, that is neural networks are prone to overfitting. Many studies have shown that over-sized deep neural networks with regard to dataset size form redundant filters, which is closely related to overfitting. These filters add very little to none to the overall information encoded in the network, due to the fact that they are either similar with small variations or are shifted versions of each other \cite{rodriguez2016regularizing, dundar2015convolutional, ayinde2017nonredundant}. Overfitting is a major issue, which is why deep learning models require huge amounts of data in order to reach a decent performance. A higher amount of input data and thereby a tighter density of the empirical training distribution, leads to a lower risk of overfitting. The lack of big datasets in many real world domains prevents a majority of deep learning applications from being realized. That is why developing models, which perform well on small amounts of data out of the box, without the requirement of transfer learning or similar methods, is key for further boosting deep learning applications. 

Besides developing highly generalized deep learning networks, gaining insights into what a neural network learns and how it arrives at its decisions has increasingly become a focus area in research. For this purpose, attribution maps have been developed that visualize which input features are most important for network predictions \cite{simonyan2013deep,smilkov2017smoothgrad,zeiler2014visualizing,springenberg2014striving,bach2015pixel,montavon2017explaining, shrikumar2017learning, lundberg2017unified}.

\begin{figure*}[t]
	\centering
    \begin{subfigure}[t]{0.501\textwidth}
        \centering
        \includegraphics[width=\textwidth]{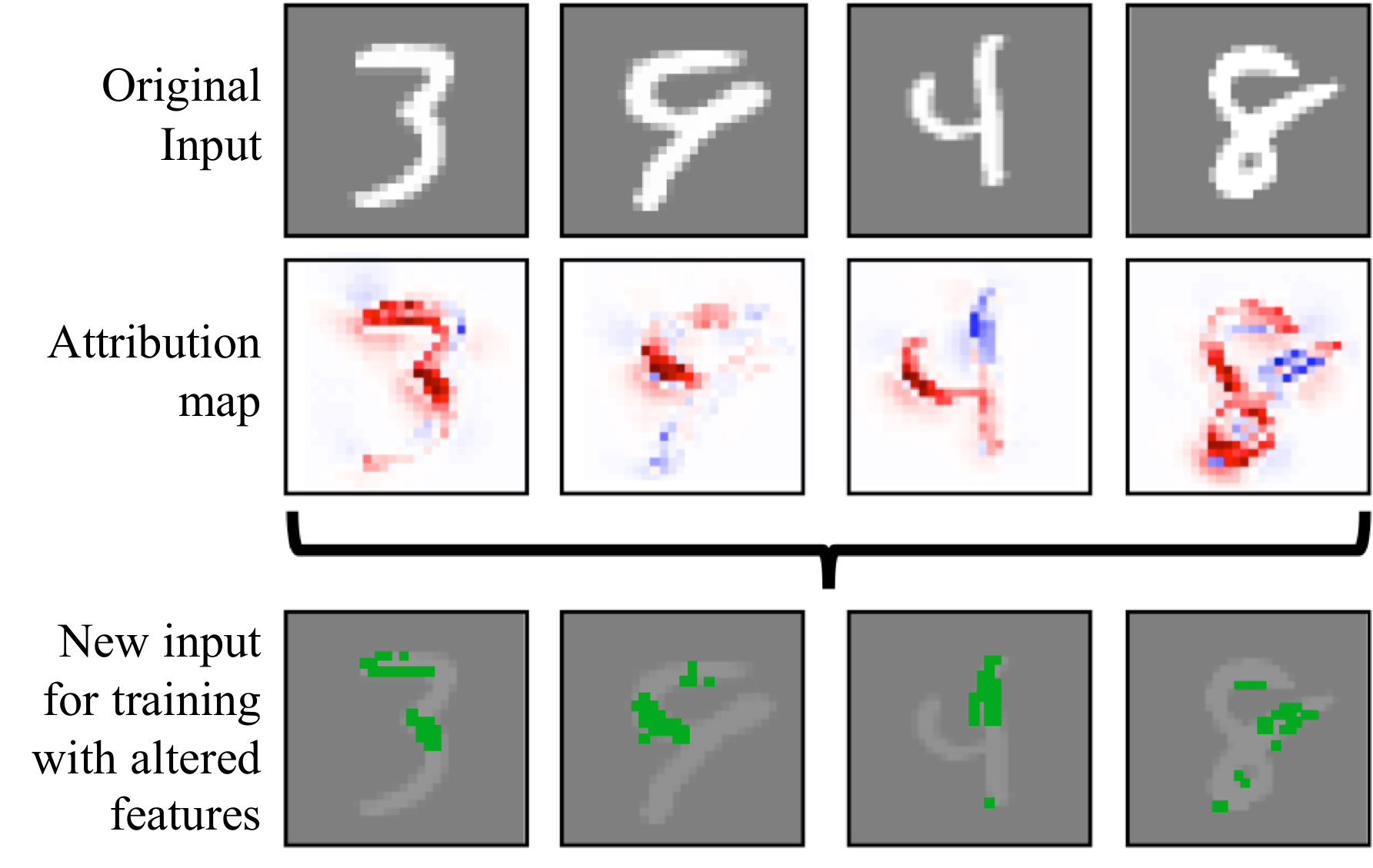}
        \caption{Challenger: Manipulation of input}
    \end{subfigure}
    \begin{subfigure}[t]{0.16\textwidth}
		\centering
    	\includegraphics[width=\textwidth]{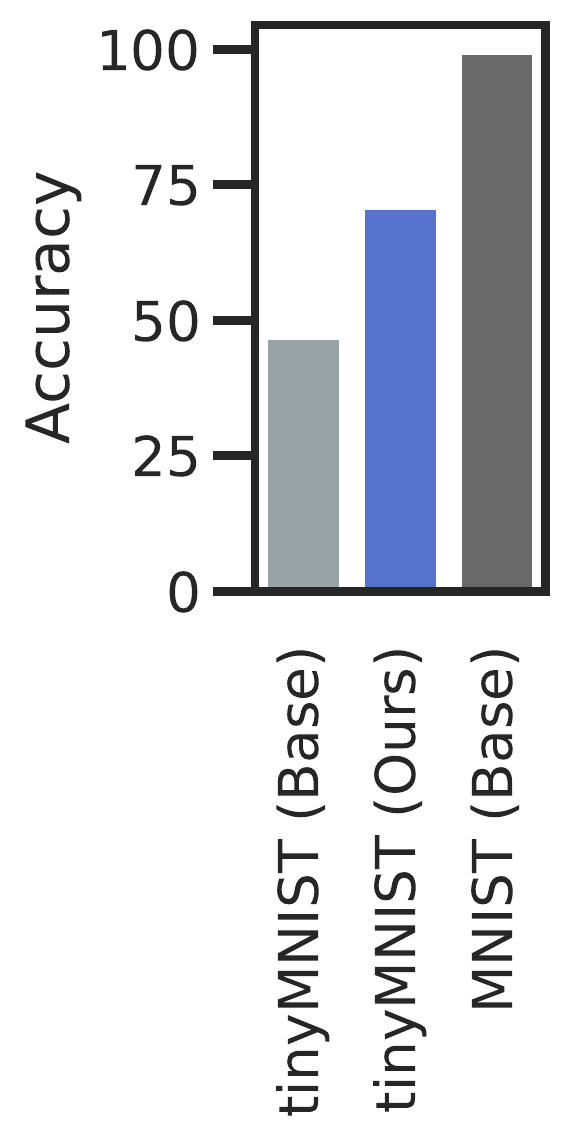}
		\caption{Accuracy}
	\end{subfigure}
	\begin{subfigure}[t]{0.174\textwidth}
		\centering
		\includegraphics[width=\textwidth]{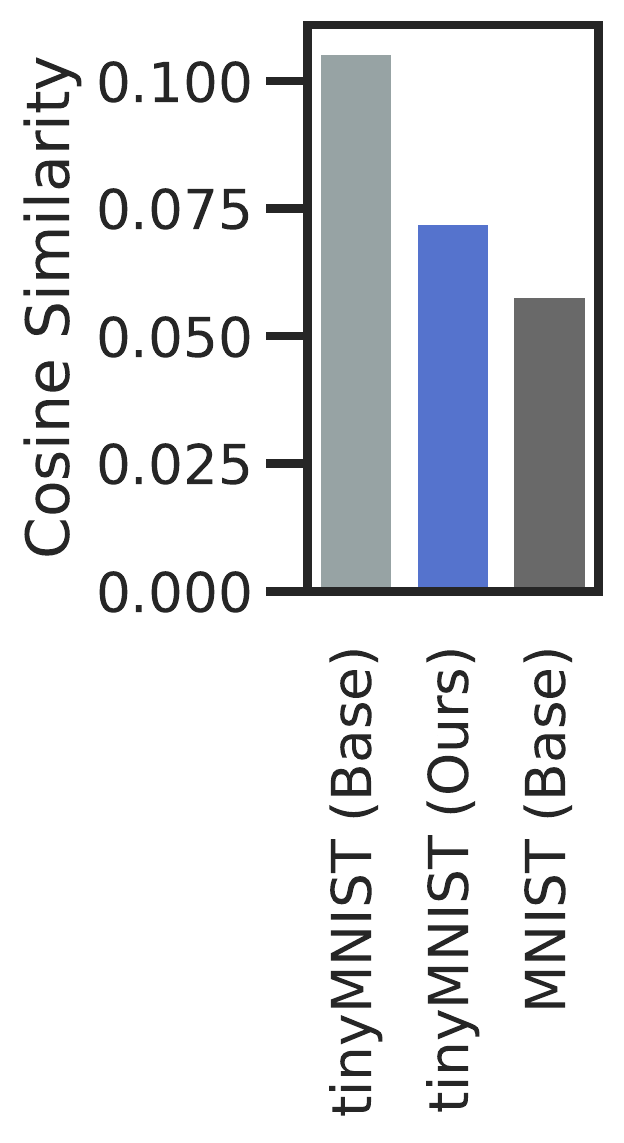}
		\caption{Cos. Similarity} 
	\end{subfigure}
	\caption{Leveraging attribution maps for training through altering input features based on their relevance regarding predictions increases performance and improves diversity among filters. We use a subset of MNIST (tinyMNIST) with only 30 samples and show the regularization performance of our Challenger training approach ("Ours") compared to state-of-the-art training ("Base"). (a) The proposed Challenger identifies features based on positive and negative relevance and alters them (green colored pixels). (b) Using our approach, the performance on tinyMNIST improves by more than half compared to state-of-the-art training, resulting in an accuracy of 71\%. This is about half the way to the performance of a model trained on the original MNIST dataset with only 0.05\% of data samples. (c) The cosine similarity between filters, a measure for filter diversity, improves (=decreases) by 33\% with our approach on tinyMNIST and is close to that of a model trained on original MNIST.}
	
\label{fig:teaser}
\end{figure*}

Neural network models are prone to incorrectly identifying in-domain test samples, which live in low-density regions of the input data manifold. This is naturally a substantial problem in datasets with few samples and a high amount of features, such as image, language (with high dimensional embedding spaces) or long multivariate time series datasets. Nevertheless, this issue is distinct from the field of out-of-distribution (OOD) samples, where a rich body of research exists already \cite{thulasidasan2019mixup, tomani2020post, snoek2019can, tomani2021parameterized, zhang2020mix, tomani2021towards}. The difference to in-domain scenarios is that OOD samples are considered to not be lying on the input data manifold. 

Enriching the empirical distribution of training data is of tremendous importance for increasing the expressive power of neural networks, especially when training models on little data. We argue that deep learning models inherently carry information about which input features to perturb in order to expose \textit{regions of ambiguity} towards separating classes on the ground-truth data manifold. Our goal is it to reduce these ambiguous regions by building new as well as more diverse filters within the network during training. We introduce a method, which utilizes the explainable power of attribution maps for training neural networks. By considering the relevance of input features with respect to predictions, we are able to grasp which features the network relies on for a certain prediction and alter these respective input patterns accordingly. We call our method "\textit{Challenger}", and we demonstrate that this method results in substantially better performance on a wide range of data domains.

\subsection{Contributions}
\begin{itemize}
    \item We introduce a generic as well as domain-independent yet effective method for training neural networks, which leverages attribution maps and is particularly suited for little data as well as consistently increases performance on big datasets.
    \item We demonstrate that utilizing attribution maps offers great potential for regularizing networks and that the key to performance improvements of the proposed approach is in fact feature selection based on attribution methods.
    \item We show that our approach yields state-of-the-art results regarding prediction performance as well as uncertainty calibration on a variety of datasets from 3 different domains: vision, natural language and time series.
\end{itemize}

\section{Related work}

In this section, we review common attribution methods and existing techniques towards regularizing neural networks. 

\subsection{Attribution methods}

Besides building high performing deep learning networks, developing insights into how such models make decisions has gained a lot of attention in recent years. The main question behind these approaches is, which input features are most important for the network to make a decision. Simonyan et al. \cite{simonyan2013deep} calculated a saliency map by maximizing the class score with regard to the image and Smilkov et al. \cite{smilkov2017smoothgrad} take this approach further by reducing noise. Further improvements of gradient based approaches include deconvolution \cite{zeiler2014visualizing} and guided backpropagation \cite{springenberg2014striving}. Bach et al. \cite{bach2015pixel} introduced the concept of conservation of total relevance, which has been adopted by others as well. Their method, called Layer-wise Relevance Propagation (LRP), backpropagates relevance (=attribution) from output to the input and forms a relevance map. Decomposition \cite{montavon2017explaining} extends LRP by decomposing activations according to contributions from its inputs and DeepLift improves backpropagating the relevance \cite{shrikumar2017learning}. Lundberg et al. \cite{lundberg2017unified} introduced Shapley Additive explanations (SHAP) based on approximating the shapley value.

\subsection{Regularization methods}

Regularization methods can broadly be divided into internal (model side) and external methods (data side). Internal methods span from well-known methods such as weight decay, which penalize complexity and special forms of weight initialization \cite{he2015delving}, to dropout \cite{srivastava2014dropout}, which prevents co-adaptations by randomly dropping nodes. Batchnorm \cite{ioffe2015batch} is another methods, which focuses on minimizing covariance shift. Furthermore, penalizing low entropy output distributions has been shown to be effective for regularization \cite{pereyra2017regularizing}. 
Researchers have demonstrated that many overfitted networks rely on redundant filters, which add little to no information to feature hierarchy \cite{rodriguez2016regularizing, dundar2015convolutional}. In order to mitigate shifted or similar versions of features, Lubana et al. \cite{lubana2020orthoreg} decorrelate filters by locally enforcing feature orthogonality and Ayinde et al. \cite{ayinde2019regularizing} maximize the encoded information in features by penalizing low cosine distances between filters. 

Another field of research focuses on adversarial training, which depending on the method, can be classified as either model and/or data sided. The fast gradient method (FGSM) adds noise to input data whose direction is the same as the gradient of the cost function with respect to the data \cite{goodfellow2014explaining}. A more powerful method to generate adversarial examples is a multi-step variant of FGSM, which applies projected gradient descent (PGD) several times \cite{kurakin2016adversarial}. A problem with adversarial training, which has been widely studied in the literature, is the trade-off between robustness and accuracy. Even though adversarial training can be effective in reducing vulnerability to such attacks, these modified training methods often decrease accuracy on natural unperturbed inputs \cite{zhang2019theoretically, raghunathan2019adversarial, madry2017towards}.

Regularization methods that focus on data involve augmentation techniques, which in case of images include: geometric transformations, flipping data features, cropping, rotation, translation and color space transformations \cite{shorten2019survey}. In order to generate meaningful transformations with these augmentation techniques, the data distribution has to be well understood by humans. In addition, Generative Adversarial Networks (GANs) are used to generate additional input data \cite{bowles2018gan} and Random Forest models are used for feature importance ordering to adjust the level of noise accordingly \cite{moreno2018forward}. Both aforementioned approaches require an additional model. Masking input data is another way, which has successfully been applied for self-supervised pretraining \cite{devlin2018bert, bao2021beit}. MixUp \cite{zhang2017mixup, thulasidasan2019mixup} is a method, which trains a neural network on convex combinations of pairs of examples and their labels; thus, favoring simple linear behavior in-between training examples. 

However, none of these methods take LRP based attribution maps into account for selecting relevant features and leverage this information for perturbing input data to regularize neural networks.

\section{Method description}

\subsection{Definitions and problem set-up}

The proposed method consists of two parts. At first, Layer-wise Relevance Propagation (LRP) \cite{bach2015pixel} is used in order to identify input features that yield positive or negative relevance for network predictions. Secondly, a \textit{Challenger} module alters input data features according to their relevance and applies a particular perturbation paradigm to challenge the underlying neural network. 

\subsection{Attribution module}
We use LRP proposed by Bach et al. \cite{bach2015pixel}, which is based on the idea of feature-wise decomposition. LRP is an established well known method that has been proofed to give sufficiently qualitative relevance maps. Let $X \in \mathbb{R}^D$ be the $D$-dimensional input and $Y \in {1,...,C}$  be the labels in a classification task with $C$ classes. Note that we do not make any assumptions about the input distribution when applying our Challenger module, that is the dependency of color channels per pixel or the embedding dimensions per word are disregarded, we even flatten the input data structure and rely solely on the relevance of each input feature independent of its dependencies to other features. That makes our method a generic method, which can be applied to any dataset. This does not affect the input structure to the underlying neural network. 

The goal of LRP is to understand the contribution $R_i$ of a single input feature $X_i$ to the prediction $f(X)$ with a mapping of $f: \mathbb{R}^{D} \rightarrow \mathbb{R}^{1}$ such that $f(X)>0$:

\begin{equation}
\label{eqn:conserv}
   f(x)=\sum_{i=1}^{D}  R_i
\end{equation}

where $i \in D$ denotes the index of an input feature $X_i$ in $X$ with its corresponding relevance $R_i$. The qualitative interpretation of relevance $R_i$ is that a feature with $R_i<0$ contributes evidence against the prediction function $f(x)$ and $R_i>0$ contributes evidence for $f(X)$.

The propagation rule is subject to the conservation property as given by Equation \ref{eqn:conserv} and essentially states that all relevance $R_k$ that has been received by a neuron is redistributed to the lower layer $l-1$ in equal amount:  

\begin{equation}
    R_j = \sum_{k}^{} \frac{z_{jk}}{\sum_{j}z_{jk}}R_k
\end{equation}

where $z_{jk}$ denotes the amount to which neuron $j$ in the preceding layer $l$ has contributed to the relevance of neuron $k$ in the subsequent layer. This equation can be reformulated by replacing $z_{jk}$ by activations $a{j}$ and weights $w_{jk}$: 

\begin{equation}
    R_j = \sum_{k}^{} \frac{a_j w_{jk}}{\epsilon+\sum_{j}a_j w_{jk}}R_k
\end{equation}

We use the $\epsilon$-rule since it is close to the generic basic rule, yet it filters out gradient noise. By adding a small positive term $\epsilon$ to the denominator, weak relevance scores are deminished and strong ones are preserved. Hence, sparser explanations regarding input features with less noise are achieved. Due to the fact that our subsequent Challenger module is mostly interested in strong positive and negative relevance, this $\epsilon$-rule is a reasonable choice.  

\subsection{Challenger module}

\subsubsection{Motivation}

Learning relevant patterns is important for neural networks in order to make correct predictions. Deemphasizing (\textit{Challenge A}) and Emphasizing (\textit{Challenge B}) the most relevant input patterns, whether they contribute positively or negatively to the prediction, encourages the network to more robustly learn these patterns and reduce ambigious regions. Perturbing input features in a directed way, regarding their relevance, increases the density of the empirical input distribution and thus contributes to an increase in diversity of filters within the network.

\subsubsection{Types of challenges}

Input data can be modified by 2 challenges $g_A:\mathbb{R}^{D} \rightarrow \mathbb{R}^{D}$ and $g_B:\mathbb{R}^{D} \rightarrow \mathbb{R}^{D}$ with set $M \subset X$ containing features that need to be modified: \\
\newpage 
\textbf{Challenge A:} Decrease a certain subset of input features $X_i$ of $X$ with respect to $M$ by a predefined value $\alpha$: \\

\begin{equation}
    g_A(X_i,M,\alpha) = 
\begin{cases}
    X_i-\alpha,& \text{if } X_i \in M\\
    X_i,              & \text{otherwise}
\end{cases}
\end{equation}

\textbf{Challenge B:} Increase a certain subset of input features $X_i$ of $X$ with respect to $M$ by a predefined value $\beta$: \\

\begin{equation}
    g_B(X_i,M,\beta) = 
\begin{cases}
    X_i+\beta,& \text{if } X_i \in M\\
    X_i,              & \text{otherwise}
\end{cases}
\end{equation}

\subsubsection{Choosing input features for modification}

Input features that are modified by one of the above challenges are selected according to the relevance maps of the prior LRP module. There are 2 types of classes the LRP module can use to backpropagate the relevance from. Either it uses the target class or a non-target class. The classes, which the Challenger can randomly choose from, are given by $K$, which indicates the top-k classes ($K$ classes with the highest predictions for the respective sample). Once the attribution map has been calculated the features are ranked according to their relevance, and either the $N$ highest or $N$ lowest feature values per sample are added to set $M$ of input features used for modification. Whether the highest or lowest features are added to set $M$ is randomly chosen for every sample. 

\subsubsection{Algorithm}

For each training batch a subset $X_A$ of 25\% of samples is modified by \textit{Challenge A}, where samples are transformed by $\tilde{X}_A=g_A(X_A,M,\alpha)$, further 25\% ($X_B$) are modified by \textit{Challenge B} with $\tilde{X}_B=g_B(X_B,M,\beta)$ and the remaining 50\% ($X_O$) are pure training samples $\tilde{X_O}=X_O$ that are not being modified in any way. We choose to keep half of unmodified samples in the training batch, because this ensures that the network does not shift away from the training distribution, but instead leverages the modified samples as a mean to first of all enrich the empirical training distribution and second of all us it as a challenge to incorporate richer information into the network during training and building unique receptive fields that extract distinct features from input data. The exemplary training procedure can be seen in Algorithm 1. 

\begin{figure*}[t]
	\begin{minipage}{\textwidth}
        \begin{algorithm}[H]
        	\caption{Training with Challenger\\ \textbf{Input}: Classifier $Y=f(\theta,X)$, attribution map generator $R=h(f(\theta,X),Y,k)$ training set $(X, Y)$, model parameters $\theta$, number of training steps $S$, number of top-k classes $K$ for calculating attribution maps, values $\alpha$ and $\beta$ for decreasing and increasing input features, number of features $N$ to be modified.}
        	\begin{algorithmic}[1]
        	\FOR{$\varsigma$ in 1:$S$}
            \STATE Read batch $B = (X_B,Y_B) = (\{X_B^{(1)} ,\dots, X_B^{(\beta)}\}, \{Y_B^{(1)} ,\dots, Y_B^{(\beta)}\})$ from training set
            \STATE Compute forward pass based on $B$: $\hat{Y}_B=f(\theta,X_B)$
            \STATE Choose $k$ randomly from $K$ and calculate attribution scores for batch with respect to the $k$-th class $R_B=h(f(\theta,X_B),\hat{Y}_B,k)$
            \STATE Build subset $M_B \in X_B$ by selecting either $N$ highest or $N$ lowest features per sample based on $R_B$
            \STATE Apply challenges to input batch $X_B$ by transforming it to $\tilde{X_B}$ based on $X_B$ and $M_B$: \\
                Split $X_B$ into chunks of $25\%$ / $25\%$ / $50\%$: $X_{B0.25,A}$ / $X_{B0.25,B}$ / $X_{B0.50,O}$\\
                $25\%$ of samples in batch: Challenge A: $\tilde{X}_{B0.25,A} = g_A(X_{B0.25,A},M_B,\alpha)$ \\
                $25\%$ of samples in batch: Challenge B: $\tilde{X}_{B0.25,B} = g_B(X_{B0.25,B},M_B,\beta)$ \\
                $50\%$ of samples in batch: Data as it is: $\tilde{X}_{B0.50,O} = X_{B0.50,O}$  \\
                Concatenate subsets: $\tilde{X_B} = (\tilde{X}_{B0.25,A}, \tilde{X}_{B0.25,B}, \tilde{X}_{B0.50,O})$
            \STATE Compute forward pass based on $\tilde{B}$:  $\hat{\tilde{Y}}_B=f(\theta,\tilde{X}_{B})$
            \STATE Compute $L_\theta$ based on $\tilde{B}$ and do one training step optimizing $\theta$
            \ENDFOR
            \end{algorithmic}
        \end{algorithm}
	\end{minipage}
\end{figure*}

\section{Experiments and results}

We quantify the performance of our Challenger module based on various vision, natural language and time series datasets. In particular, we use MNIST \cite{lecun2010mnist}, CIFAR10 \cite{krizhevsky2009learning}, Reuters \cite{duauci2019}, 20Newsgroups \cite{duauci2019} and PTB-XL \cite{wagner2020ptb}, each of them with training, validation and test splits. Both vision datasets, namely MNIST and CIFAR10 have 10 classes, whereas MNIST is a gray-scale image dataset and CIFAR10 is a color image dataset with 3 input channels (RGB) per pixel. Reuters and 20Newsgroups are datasets, where news articles are modelled as sequences of words with the objective of correctly classifying the respective category. Additionally to the original datasets, we produced small datasets with a lot less samples 
in order to analyze the behavior of our Challenger module also on very small datasets. We use the prefix \textit{"tiny"} for these datasets in the following.  

For time series classification we use a special dataset, particularly the PTB-XL \cite{wagner2020ptb}, which is the to-date largest publicly accessible clinical ECG dataset in the field of Electrocardiography (ECG). ECG is a key diagnostic tool to assess the cardiac condition of a patient. The dataset contains 71 different statements conform to the SCP-ECG standard and covers diagnostic, form and rhythm statements. In our experiments we follow the recommended train-validation-test splits of Wagner et al. \cite{wagner2020ptb} and use a sampling frequency of 100Hz. The dataset comprises 21837 clinical 12-lead ECG records from 18885 patients of 10 seconds length.

For each dataset we use an appropriate state-of-the-art classifier. For MNIST we use VGG11 \cite{simonyan2014very}, 
for CIFAR10 we use VGG16, for Reuters as well as 20Newsgroups we use a CNN with 3 convolutional layers and 2 linear layers. In the case of both natural language datasets we generate vector representations of words using the pre-trained GLOVE embedding \cite{pennington2014glove} (length 100). For PTB-XL we use the fully convolutional network (FCN) from \cite{wang2017time}. We use Adam optimizer without learning rate scheduling to optimize cross entropy loss. Attribution maps are always calculated for inputs and in the case of Reuters and 20Newsgroups we process attribution maps for embeddings and alter them with Challenger, respectively. 

In contrast to tiny datasets, where we use Challenger throughout the whole training procedure, we demonstrate on our standard datasets that our approach can also be used for finetuning. We take the fully trained classifiers and finetune them with our Challenger module. Note that in contrast to traditional finetuning, do not freeze any layers and we keep the learning rate the same as it was during original training to avoid any additional regularizing effects.

The results section is structured as follows: \textbf{1.} We analyze our Challenger module on different scenarios, namely on all 3 data domains, with different sample sizes and different models: tinyMNIST (with unregularized classifier and Challenger applied troughout training), REUTERS (with unregularized classifier and Challenger finetuning) and PTB-XL (with regularized classifier (dropout and weight decay) and Challenger finetuning). \textbf{2.} We investigate the behavior of Challenger on tiny datasets and standard datasets with completely unregularized models. In order to make sure that we measure the actual performance gain of our proposed model and that performance gains from established regularization methods do not factor into the result of our Challenger module.
\textbf{3.} We apply a classifier on PTB-XL with state-of-the-art regularization methods in order to investigate the additional gains of our proposed training method. 

\subsection{Analysis of our Challenger module}

\subsubsection{Diverse filters}

\begin{figure*}
	\centering
    \begin{subfigure}[t]{0.3\textwidth}
        \centering
        \includegraphics[width=\textwidth]{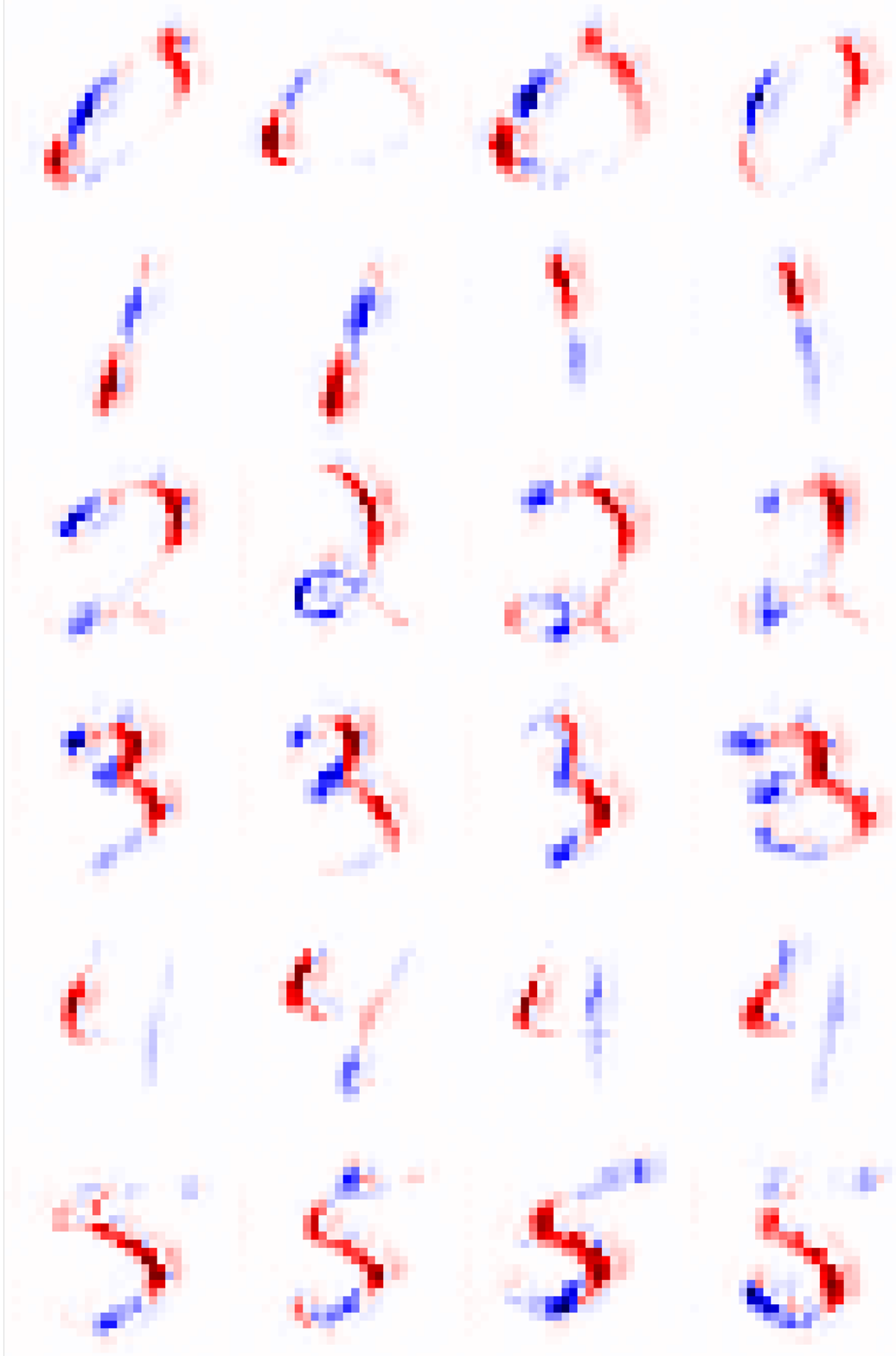}
        \caption{tinyMNIST Base}
    \end{subfigure}
    \begin{subfigure}[t]{0.3\textwidth}
		\centering
    	\includegraphics[width=\textwidth]{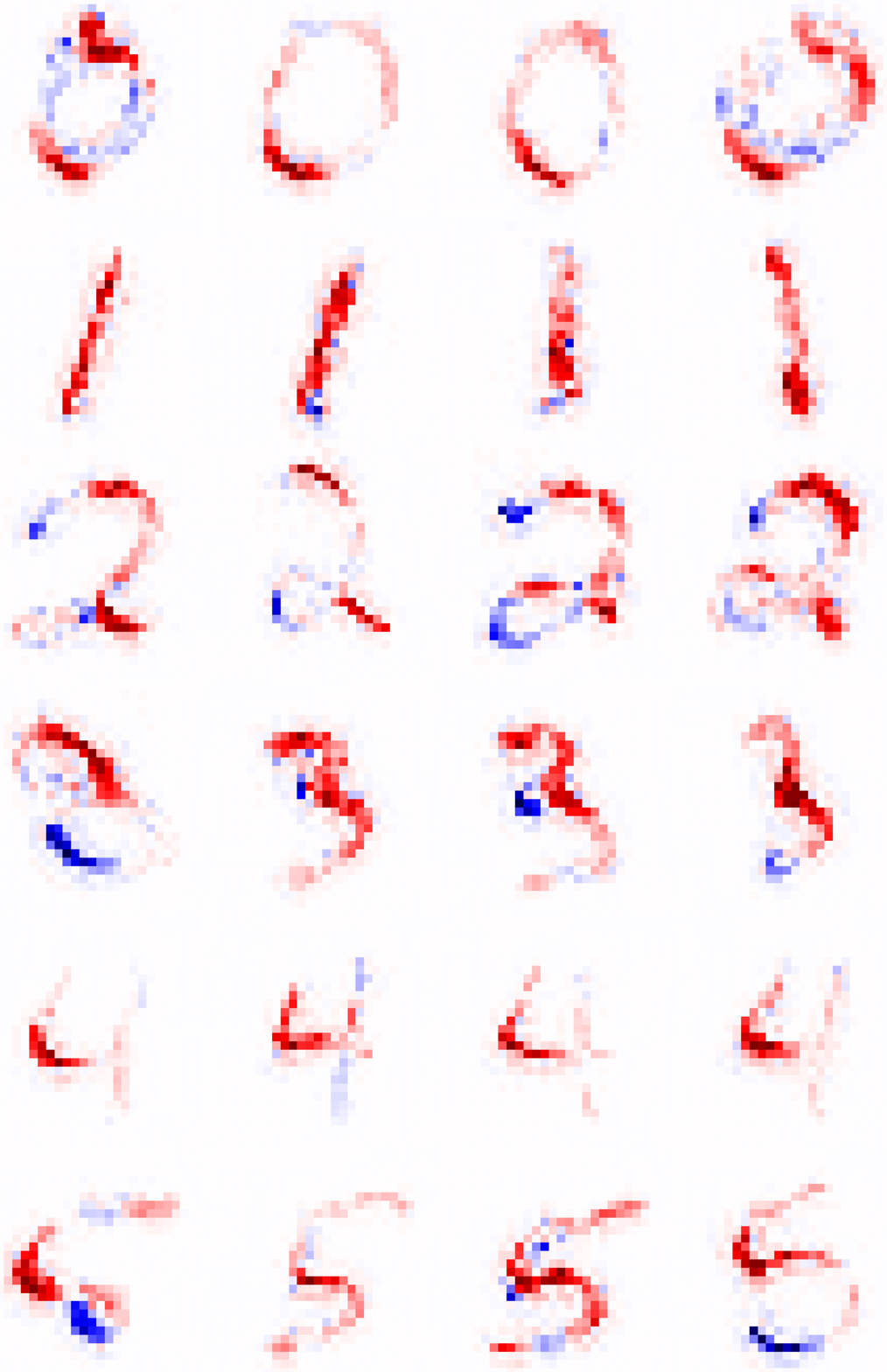}
		\caption{tinyMNIST Challenger}
	\end{subfigure}
	\begin{subfigure}[t]{0.3\textwidth}
		\centering
		\includegraphics[width=\textwidth]{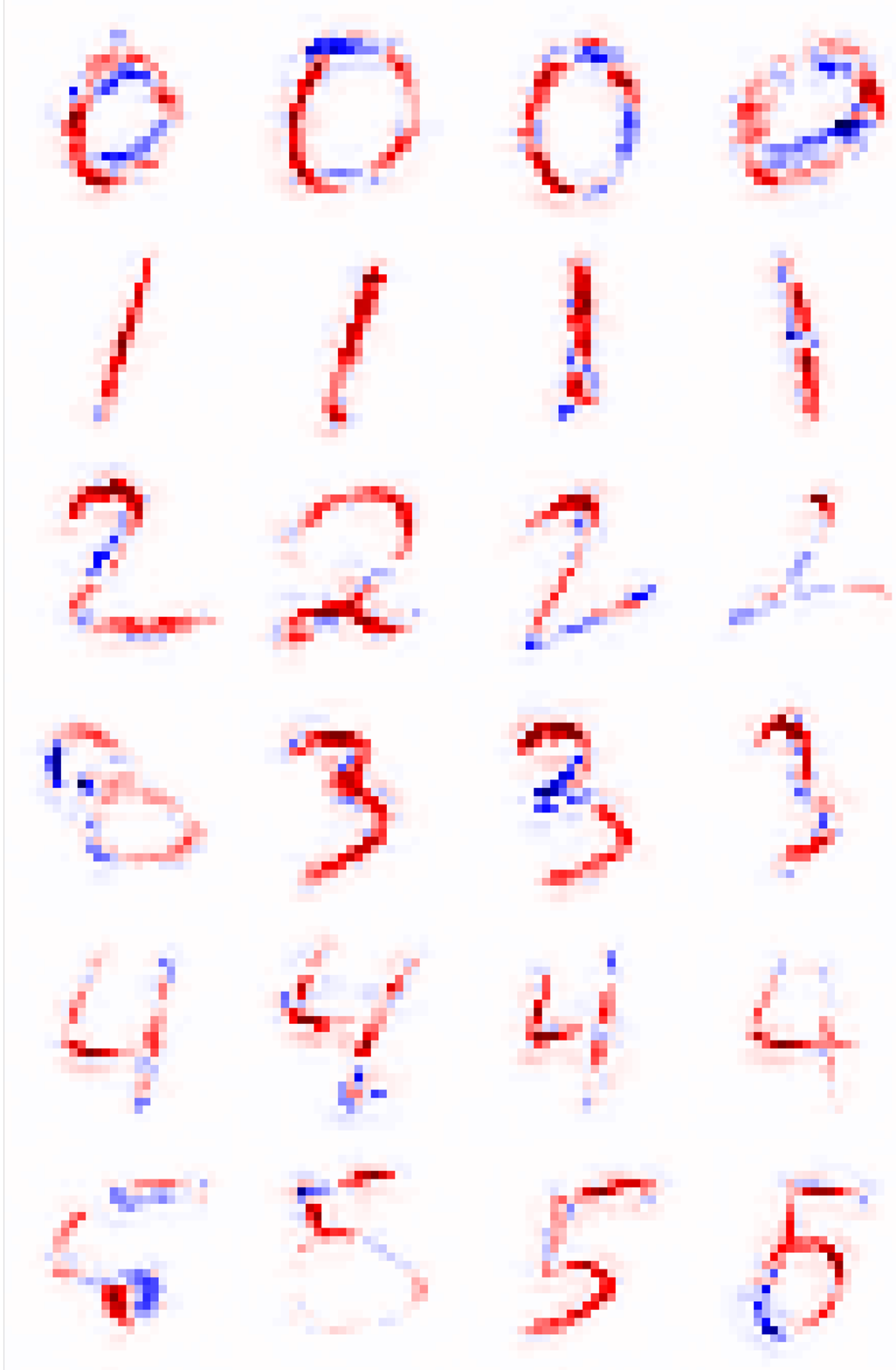}
		\caption{MNISTbig Base}
	\end{subfigure}
	\caption{Attribution maps for VGG11 traind with state-of-the-art training (Base) and our Challenger training module (Challenger). Red indicates positive relevance and blue negative relevance with respect to the target class).}
\label{fig:attribution_maps}
\end{figure*}

As a first step in our analysis we want to shed light onto the inner workings of a neural network trained with our Challenger module. First we show attribution maps derived with LRP from a state-of-the-art VGG11 model on tinyMNIST with only 30 samples (3 per class) compared to standard MNIST with 60000 (minus validation set) (Figure \ref{fig:attribution_maps}(a)). For tiny MNIST these attribution maps look very similar within the same classes with respect to positive (red) and negative (blue) relevance. This indicates a limited diversity of filters and hence a worse adaptiveness to slightly different test data samples. To test this assumption of non-diverse features we make use of a common correlation measure, namely the cosine similarity between filters of each layer and average the mean cosine similarity of every layer in Table \ref{tab:abl_metrics}. The cosine similarity is nearly twice as high for tinyMNIST as it is for regular MNIST, both trained with an unregularized network (low cosine similarity indicates more diverse filters). Challenger on the other hand achieves a mean cosine similarity score of 0.0724 on tinyMNIST, which is close to that of MNIST with 0.058. This behavior reflects also on attribution maps, where tinyMNIST with only a few samples trained with Challenger expresses much more diverse attribution maps (Figure \ref{fig:attribution_maps}(b)) than without Challenger (Figure \ref{fig:attribution_maps}(a)), and is similar to standard MNIST trained with the ordinary state-of-the-art method without Challenger, but with thousands of training samples (Figure \ref{fig:attribution_maps}(c)). Not only filters are more diverse when applying Challenger module, also accuracy improves as observed in Figure \ref{fig:teaser}. One can assume that more complex filters are better suited to correctly classify test samples that lie in ambiguous regions, which is particularly important in scenarios with few data samples.  

\begin{table*}[h]
\centering
\caption{Mean Cosine similarity (CS) between filters across layers (L1, L2, L3, etc.) and accuracy of basic training (Base) and Challenger module (Ours) on tinyMNIST and MNIST (standard deviation).}
\begin{tabular}{lccccc}
\toprule
{} & Accuracy & CS Mean & CS L1 & CS L2 & CS L3\\
\midrule
MNIST-T Base & 46.99 & 0.106 & \textbf{0.019(0.337)} & 0.191(0.130) & 0.089(0.184) \\
MNIST-T Ours & \textbf{71.06} & \textbf{0.0724} & 0.022(0.335) & \textbf{0.107(0.140)} & \textbf{0.043(0.117)} \\
\midrule
\itshape{MNIST-B Base} & \itshape{99.49} & \itshape{0.058} & \itshape{0.019(0.332)} & \itshape{0.071(0.164)} & \itshape{0.019(0.127)} \\
\bottomrule
\end{tabular}
\label{tab:abl_metrics}
\end{table*}

\subsubsection{Benefit of attribution maps}

\begin{figure*}
	\centering
    \begin{subfigure}[t]{0.25\textwidth}
        \centering
        \includegraphics[width=\textwidth]{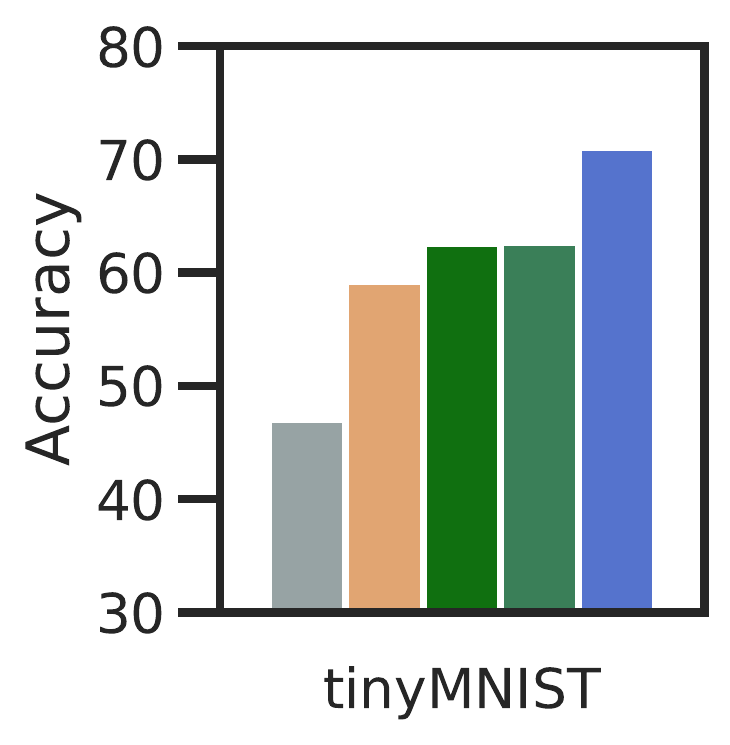}
    \end{subfigure}
    \begin{subfigure}[t]{0.25\textwidth}
        \centering
        \includegraphics[width=\textwidth]{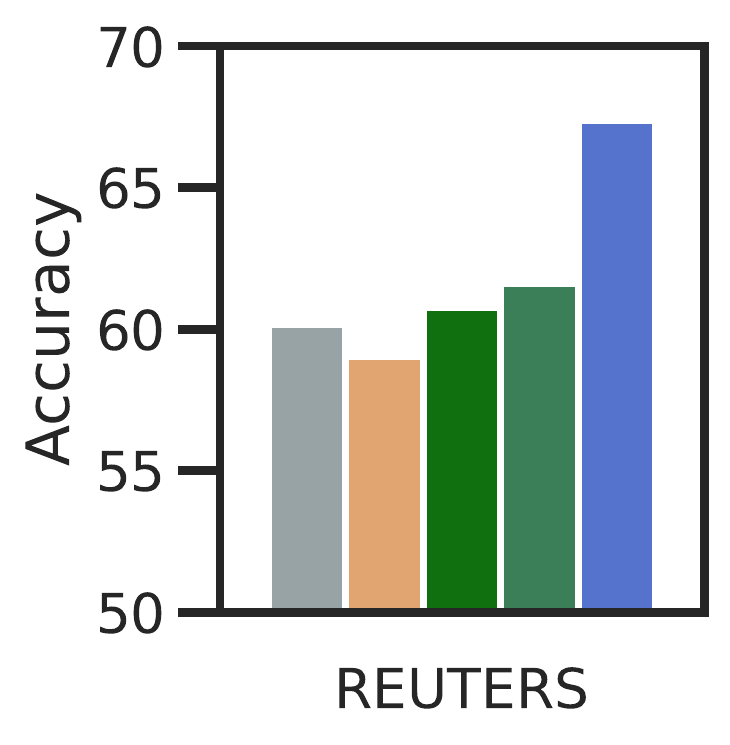}
    \end{subfigure}
    \begin{subfigure}[t]{0.44\textwidth}
		\centering
    	\includegraphics[width=\textwidth]{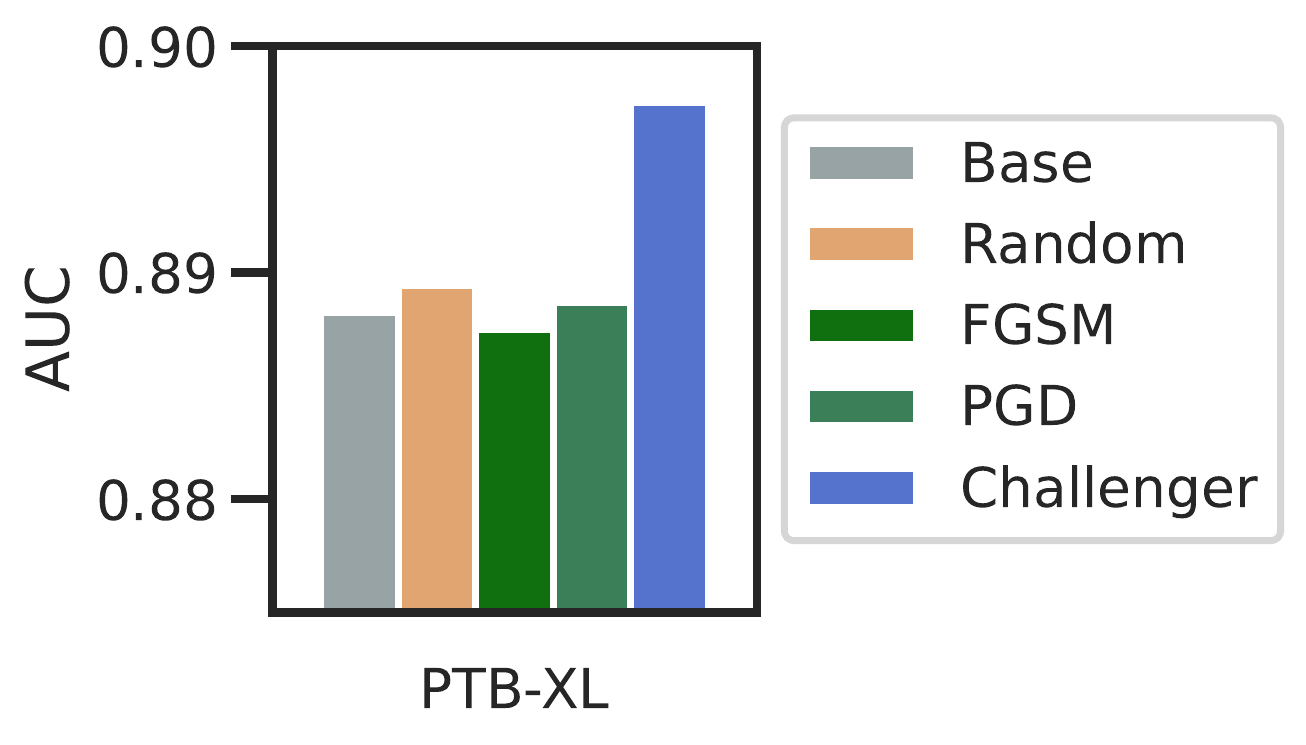}
	\end{subfigure}
	\caption{Comparison between Challenger module (selects input features according to relevance) 
	and Random (selects features randomly) as well as adversarial training methods (FGSM and PGD).}
\label{fig:abl_random}
\end{figure*}

In this section we want to investigate the benefit of using attribution maps for selecting features. In order to quantify the advantage of selecting features based on relevance scores from LRP, we replace the attribution model, which chooses the features in subset $M$, with a random generator. Essentially subset $M$, which contains the features that are beeing manipulated by our proposed challenges, is generated randomly. Furthermore, we compare Challenger with two of the most prominent adversarial training methods, namely FGSM \cite{goodfellow2014explaining} and PGD \cite{kurakin2016adversarial}. Figure \ref{fig:abl_random} indicates that randomly choosing features that are being altered does generally not work or improves accuracy only slightly. Moreover, FGSM and PGD perform consistently worse than Challenger module. This likely stems from a well known fact that these methods suffer a trade off between robustness and accuracy \cite{zhang2019theoretically, raghunathan2019adversarial, madry2017towards}. Challenger particularly chooses a small subset of relevant input patterns and perturbs them, which we observe has a positive effect on accuracy and the area under the receiver operating characteristic curve (AUC). 

\subsubsection{Sensitivity of challenger to top-k predictions}

\begin{figure*}
	\centering
    \begin{subfigure}[t]{0.3\textwidth}
        \centering
        \includegraphics[width=\textwidth]{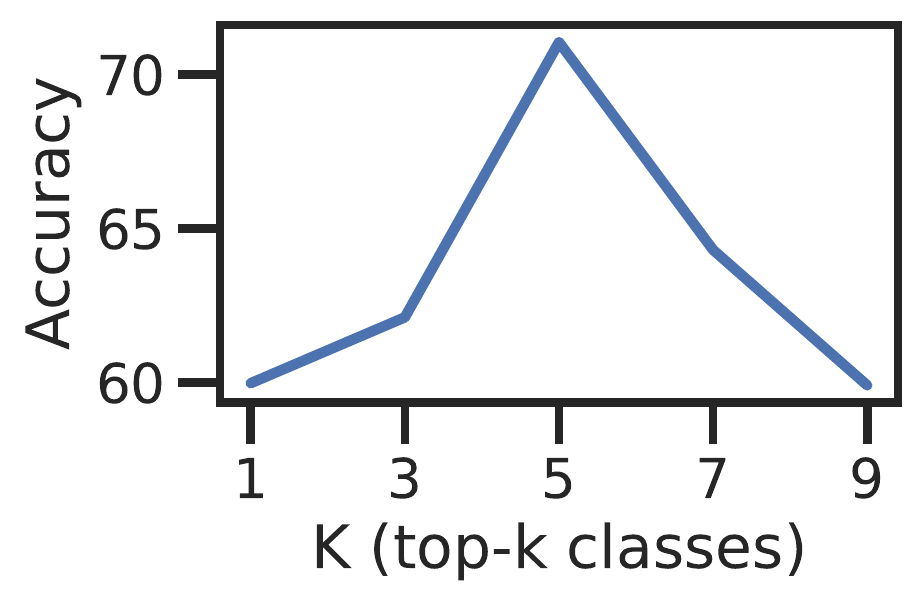}
        \caption{Vision (tinyMNIST)}
    \end{subfigure}
    \begin{subfigure}[t]{0.3\textwidth}
        \centering
        \includegraphics[width=\textwidth]{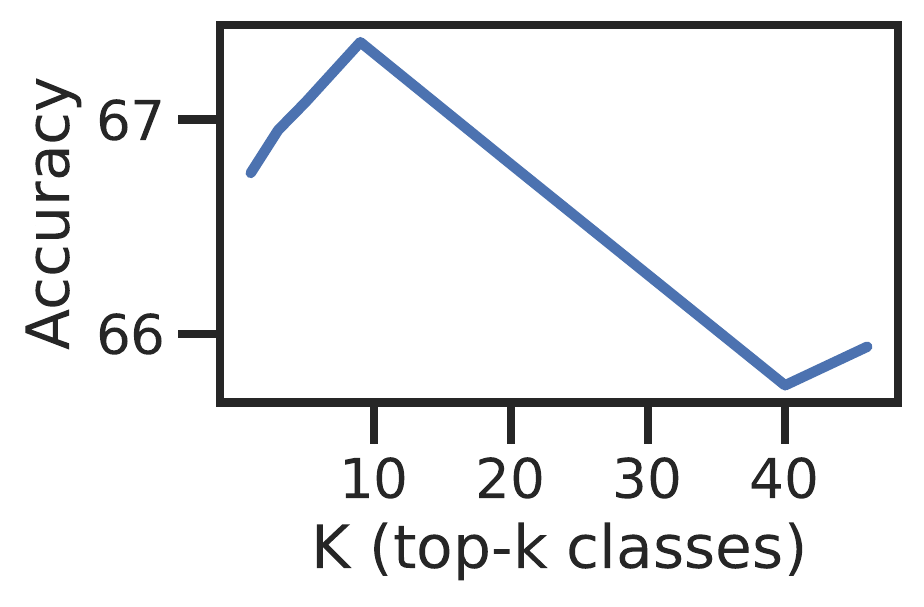}
        \caption{NLP (REUTERS)}
    \end{subfigure}
    \begin{subfigure}[t]{0.33\textwidth}
		\centering
    	\includegraphics[width=\textwidth]{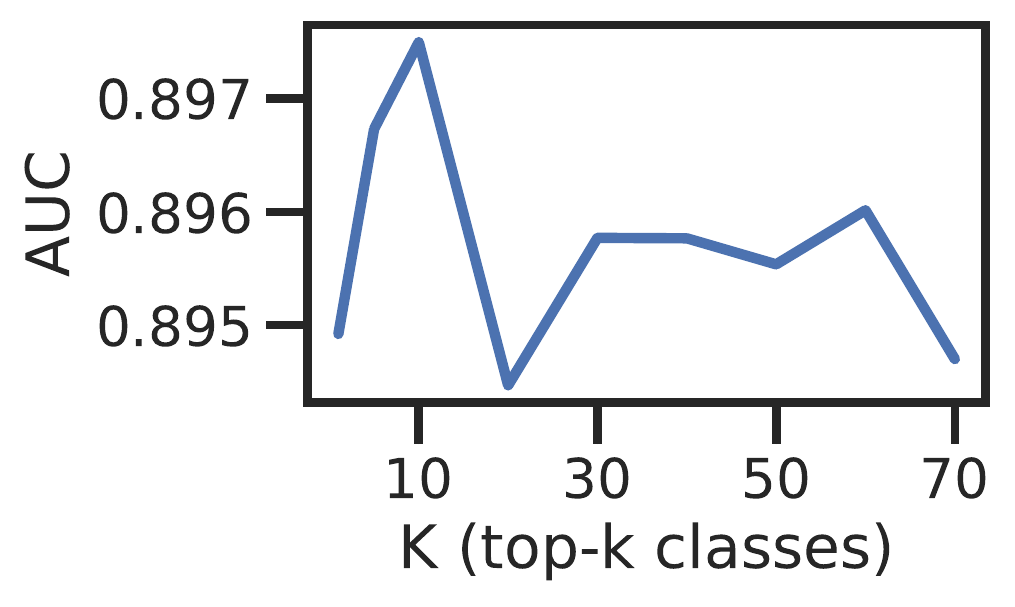}
		\caption{Time Series (PTB-XL)}
	\end{subfigure}
	\caption{Sensitivity of Challenger to top-k predictions ($K$) for attribution map generation.}
\label{fig:abl_topk}
\end{figure*}

Another contribution of the Challenger module is that not only attribution maps that are calculated with respect to target classes are utilized, but also a set of attribution maps regarding top-k (=$K$) classes are taken into account. We want to understand the sensitivity of Accuracy and AUC with respect to top-k classes. Therefore, we conducted a sensitivity analysis based on the best performing model, and alter $K$ in Figure \ref{fig:abl_topk}. In general the performance initially increases and after a peak starts to decrease. An explanation for this behavior could be that increasing $K$ initially adds information to the network until a certain point where the limit of the capacity of the network is reached. Nevertheless, the performance fluctuates only mildly with respect to $K$ and always stays clearly ahead of the basic model's performance. On the basis of these results we chose $K$=5 for our vision datasets, $K$=10 for our natural language processing datasets and $K$=10 for our time series dataset. 

\subsection{Analysis on various datasets}

In this section we apply the Challenger module to completely unregularized classifiers in order to avoid measuring external regularization effects, which do not stem from the Challenger module. First of all, we investigate tiny datasets shown in Table \ref{tab:metrics_tinydatasets}. Second of all, we show results for standard datasets with many more training examples (Table \ref{tab:metrics_datasets}). For tiny datasets and standard datasets accuracy, negative log-likelihood and brier score are almost always improved by Challenger compared to the state-of-the-art training method. Additionally, the Expected Calibration Error (ECE) is decreased by the Challenger module, which demonstrates better uncertainty calibration. Clearly Challenger performs particularly well with only little data as well as increases performance on big datasets.

\begin{table*}
\centering
\caption{Metrics for tiny datasets for Challenger (Ours) and state-of-the-art training (Base).}
\begin{tabular}{l|cc|cc|cc|cc}
\toprule
\multirow{1}{*}{Dataset} &
  \multicolumn{2}{c|}{Accuracy [\%]} &
  \multicolumn{2}{c|}{NLL} &
  \multicolumn{2}{c|}{Brier} &
  \multicolumn{2}{c}{ECE} \\
& Base & Ours & Base & Ours & Base & Ours & Base & Ours \\
\midrule
tinyMNIST & 46.99 & \textbf{71.06} & 7.670 & \textbf{3.130} & 0.096 & \textbf{0.052} & 0.454 & \textbf{0.242} \\
tinyCIFAR & 34.72 & \textbf{38.68} & 8.198 & \textbf{3.515} & 0.116 & \textbf{0.100} & 0.540 & \textbf{0.427} \\
tinyReuters & 36.9 & \textbf{45.55} & 0.027 & \textbf{0.016} & 0.018 & \textbf{0.013} & 0.160 & \textbf{0.047} \\
tiny20Newsgroups & 35.2 & \textbf{40.55} & 0.038 & \textbf{0.026} & 0.049 & \textbf{0.043} & 0.375 & \textbf{0.313} \\
\bottomrule
\end{tabular}
\label{tab:metrics_tinydatasets}

\caption{Metrics for standard datasets for Challenger (Ours) and state-of-the-art training (Base).}
\begin{tabular}{l|cc|cc|cc|cc}
\toprule
\multirow{1}{*}{Dataset} &
  \multicolumn{2}{c|}{Accuracy [\%]} &
  \multicolumn{2}{c|}{NLL} &
  \multicolumn{2}{c|}{Brier} &
  \multicolumn{2}{c}{ECE} \\
& Base & Ours & Base & Ours & Base & Ours & Base & Ours\\
\midrule
MNIST & 99.49 & \textbf{99.54} & 0.048 & \textbf{0.024} & \textbf{0.001} & \textbf{0.001} & 0.005 & \textbf{0.004}\\
CIFAR & 80.00 & \textbf{81.05} & 6.902 & \textbf{2.732} & 0.039 & \textbf{0.034} & 0.193 & \textbf{0.159} \\
Reuters & 60.15 & \textbf{67.36} & 0.022 & \textbf{0.018} & 0.011 & \textbf{0.010} & \textbf{0.115} & 0.117 \\
20Newsgroups & 70.18 & \textbf{73.85} & \textbf{0.020} & 0.024 & 0.025 & \textbf{0.023} & 0.210 & \textbf{0.207} \\
\bottomrule
\end{tabular}
\label{tab:metrics_datasets}
\end{table*}

\subsection{Analysis on PTB-XL}

Finally, we are interested, whether the Challenger module is capable of increasing performance even for already regularized networks with state-of-the-art regularization methods. To this end we conduct experiments on the to-date largest publicly accessible clinical multi-label ECG dataset and closely follow Strodthoff et al. \cite{strodthoff2020deep}. Apart from using \textit{all} categories (diag, form and rhythm) we also train models on the following subsets: \textit{diag}, \textit{sub-diag} and \textit{super-diag}, which contain different numbers of classes ranging from 5 to 71. We conduct our experiment with the fully convolutional network (FCN) proposed in \cite{wang2017time} and regularize it with weight decay as well as dropout. Furthermore, we apply mixup \cite{zhang2017mixup, thulasidasan2019mixup}, an effective data augmentation method. Our experiments demonstrate that also in these settings, Challenger is able to consistently outperform the baseline as shown in Table \ref{tab:ptb-xl}.

\begin{table*}
\centering
\caption{Comparison based on AUC [\%] between regularized models (weight decay (WD), dropout (DO) and Mixup) with and without Challenger module on PTB-XL dataset. We also indicate 95\% confidence intervals for all results, obtained via bootstrapping on the test set.}
\begin{tabular}{l|cccccc}
\toprule
Method & all & diag & sub-diag & super-diag\\
\midrule
DO \& WD (Base) & 88.83(1.0) & 89.25(1.1) & 89.16(1.4) & 90.40(0.7) \\
DO \& WD (Ours) & \textbf{89.75(0.8)} & \textbf{89.49(1.4)} & \textbf{90.06(1.7)} & \textbf{90.91(0.7)} \\
\midrule
MixUp \& DO \& WD (Base) & 89.07(1.0) & 89.82(1.3) & 89.92(1.0) & 91.21(0.6) \\
MixUp \& DO \& WD (Ours) & \textbf{89.90(1.1)} & \textbf{90.77(1.5)} & \textbf{90.08(1.6)} & \textbf{91.49(0.6)} \\
\bottomrule
\end{tabular}
\label{tab:ptb-xl}
\end{table*}

\section{Conclusion}

We present a generic and versatile approach for improving generalization performance by leveraging attribution maps for training. The proposed Challenger module performs well even with little data and is able to improve prediction performance as well as uncertainty calibration on big datasets. Challenger focuses on emphasizing and deemphasizing relevant patterns in input data samples and consequently improves regularization of neural networks. Our training approach is independent of the data domain and can be readily applied to any single- or multi-label classification task. Hence, we demonstrate that Challenger can be applied to vision, natural language processing and time series classification. In fact, the proposed method does not require any prior assumptions on the input domain, because it solely depends on the attribution method. Investigating different attribution methods and their effect on the proposed Challenger module as well as on training algorithms as a whole, likely will broaden our view on model training as well as will potentially open up plenty of new ways of training deep neural networks. In conclusion, utilizing attribution maps for training offers great potential for regularizing all sorts of neural networks, improving generalization performance and at the same time increasing uncertainty awareness. 
\section*{Acknowledgements}

This work was supported by the Munich Center for Machine Learning and has been funded by the German Federal Ministry of Education and Research (BMBF) under Grant No. 01IS18036B.

{\small
	\bibliographystyle{plain}
	\bibliography{references}
}

\end{document}